\documentclass[11pt,letterpaper]{article}
\usepackage{emnlp2017}
\usepackage{times}
\usepackage{latexsym}
\usepackage{graphicx}
\usepackage{xcolor}
\usepackage{url}
\usepackage{breakurl}
\usepackage{diagbox}
\usepackage{multirow}
\usepackage{booktabs}

\newcommand\ProductAnn[1]{\textcolor{olive}{\underline{#1}}}
\newcommand\email[1]{\fontsize{9.7}{11.5}\texttt{#1}}

\emnlpfinalcopy

\title{Identifying Products in Online Cybercrime Marketplaces:\\
         A Dataset for Fine-grained Domain Adaptation}

\author{Greg Durrett\\
UT Austin\\
\email{gdurrett@cs.utexas.edu}\\
\And
Jonathan K. Kummerfeld\\
University of Michigan\\
\email{jkummerf@umich.edu}\\
\And
Taylor Berg-Kirkpatrick\\
Carnegie Mellon University\\
\email{tberg@cs.cmu.edu}\\
\AND
Rebecca S. Portnoff\\
UC Berkeley\\
\email{rsportnoff@cs.berkeley.edu}\\
\And
Sadia Afroz\\
ICSI, UC Berkeley\\
\email{sadia@icsi.berkeley.edu}\\
\And
Damon McCoy\\
NYU\\
\email{mccoy@nyu.edu}\\
\AND
Kirill Levchenko\\
UC San Diego\\
\email{klevchen@cs.ucsd.edu}\\
\And
Vern Paxson\\
ICSI, UC Berkeley\\
\email{vern@berkeley.edu}}

\date{}

\begin{document}

\maketitle

\begin{abstract}

One weakness of machine-learned NLP models is that they typically perform poorly
on out-of-domain data. In this work, we study the task of identifying products
being bought and sold in online cybercrime forums, which exhibits particularly challenging
cross-domain effects.
We formulate a task that represents a hybrid of slot-filling information extraction and
named entity recognition and annotate data from four different forums.
Each of these forums constitutes its own ``fine-grained
domain'' in that the forums cover different market sectors with different properties, even though
all forums are in the broad domain of cybercrime. We characterize
these domain differences in the context of a learning-based system: supervised models
see decreased accuracy when applied to new forums, and standard techniques for
semi-supervised learning and domain adaptation have limited effectiveness on this data,
which suggests the need to improve these techniques.
We release a dataset of 1,938 annotated posts from across the four forums.\footnote{Dataset and code to train models available at \\\url{https://evidencebasedsecurity.org/forums/}}

\end{abstract}

\section{Introduction}

\begin{figure}[t!]
  \centering
\small

\fbox{\parbox{0.97\linewidth}{
\footnotesize
\setlength{\tabcolsep}{1pt}
\begin{tabular}{ll}
  TITLE: & [ buy ] Backconnect \ProductAnn{bot} \\
  BODY: & Looking for a solid backconnect \ProductAnn{bot} . \\
        & If you know of anyone who codes them please let \\
        & me know \\
\end{tabular}
}} \\[2pt]

(a) File 0-initiator4856 \\[10pt]

\fbox{\parbox{0.97\linewidth}{
\footnotesize
\setlength{\tabcolsep}{1pt}
\begin{tabular}{ll}
  TITLE: & Exploit \ProductAnn{cleaning} ? \\
  BODY:  & Have some Exploits i need \ProductAnn{fud} . \\
\end{tabular}
}} \\[2pt]

(b) File 0-initiator10815 \\[2pt]

\caption{\label{fig:example} Example posts and annotations from Darkode, with
  annotated product tokens underlined. The second example exhibits jargon
  (\emph{fud} means ``fully undetectable''), nouns that could be a product in
  other contexts (\emph{Exploit}), and multiple lexically-distinct descriptions
  of a single service. Note that these posts are much shorter than the average
  Darkode post (61.5 words).
}
\end{figure}

\begin{table*}[t]
  \setlength{\tabcolsep}{4.9pt}
  \begin{center}
  \begin{tabular}{r|rr|rrrrr}
    \hline
    &       & Words & Products & Annotated & Annotators & \multicolumn{2}{c}{Inter-annotator agreement} \\
    Forum & Posts & per post & per post & posts & per post & 3-annotated & all-annotated \\
    \hline
    Darkode   & 3,368 & 61.5 & 3.2 & 660/100/100 & 3/8/8 & 0.62 & 0.66 \\
    Hack Forums & 51,271 & 58.9 & 2.2 & 758/140 & 3/4 & 0.58 & 0.65 \\
    Blackhat  & 167 & 174 & 3.2 & 80 & 3 & 0.66 & 0.67 \\
    Nulled  & 39,118 & 157 & 2.3 & 100 & 3 & 0.77 & - \\
    \hline
  \end{tabular}
  \end{center}
  \caption{\label{table:datasets}
	  Forum statistics.
	  The left columns (posts and words per post) are calculated over all data, while the right columns are based on annotated data only.
	  Note that products per post indicate product mentions per post, not product types.
	  Slashes indicate the train/development/test split for Darkode and train/test split for Hack Forums.
      Agreement is measured using Fleiss' Kappa; the two columns cover data where three annotators labeled each post and a subset labeled by all annotators.
  }
\end{table*}

NLP can be extremely useful for enabling scientific inquiry, helping us
to quickly and efficiently understand large corpora, gather evidence,
and test hypotheses \cite{BammanEtAl2013,OConnorEtAl2013}.
One domain for
which automated analysis is particularly useful is Internet security: researchers
obtain large amounts of text data pertinent to active threats or ongoing
cybercriminal activity, for which the ability to rapidly characterize that
text and draw conclusions can reap major benefits
\cite{krebs-target1,krebs-target2}. However, conducting automatic analysis is difficult because
this data is out-of-domain for conventional NLP models, which harms the performance of
both discrete models \cite{McCloskyEtAl2010} and deep models \cite{ZhangEtAl2017}.
Not only that, we show that data from one cybercrime forum is even out of domain with respect
to \emph{another} cybercrime forum, making this data especially challenging.

In this work, we present the task of identifying products being bought and
sold in the marketplace sections of these online cybercrime forums. 
We define a token-level annotation task where, for each post, we annotate
references to the product or products being bought or sold in that post. Having the ability to
automatically tag posts in this way lets us characterize the composition
of a forum in terms of what products it deals
with, identify trends over time, associate users with particular activity
profiles, and connect to price information to better understand the
marketplace. Some of these analyses only require post-level information
(what is the product being bought or sold in this post?) whereas other
analyses might require token-level references; we annotate at the token level
to make our annotation as general as possible. Our dataset has already proven
enabling for case studies on these particular forums \cite{PortnoffEtAl2017},
including a study of marketplace activity on bulk hacked accounts versus users selling
their own accounts.

Our task has similarities to both slot-filling information extraction (with
provenance information) as well as standard named-entity recognition
(NER). Compared to NER, our task features a higher dependence on context:
we only care about the specific product being bought or sold in a post,
not other products that might be mentioned. Moreover, because we are
operating over forums, the data is substantially messier than classical
NER corpora like CoNLL \cite{TjongKimSangDeMeulder2003}. While prior work
has dealt with these messy characteristics for
syntax \cite{KaljahiEtAl2015} and for discourse
\cite{LuiBaldwin2010,KimEtAl2010,WangEtAl2011}, our work is the first to
tackle forum data (and marketplace forums specifically) from an information
extraction perspective.

Having annotated a dataset, we examine supervised and semi-supervised learning approaches
to the product extraction problem.
Binary or CRF classification of tokens as products is effective,
but performance drops off precipitously when a system trained on one forum
is applied to a different forum: in this sense, even two different cybercrime forums
seem to represent different ``fine-grained domains.''
Since we want to avoid having to annotate data for every new forum that might need to
be analyzed, we explore several methods for adaptation, mixing
type-level annotation \cite{GarretteBaldridge2013,GarretteEtAl2013},
token-level annotation \cite{Daume2007}, and semi-supervised approaches
\cite{TurianEtAl2010,KshirsagarEtAl2015}. We find
little improvement from these methods and discuss why they fail
to have a larger impact.

Overall, our results characterize the challenges of our fine-grained
domain adaptation problem in online marketplace data. We believe that this new dataset
provides a useful testbed for additional inquiry and investigation into modeling
of fine-grained domain differences.

\section{Dataset and Annotation}

We consider several forums that vary in the nature of products being traded:
\begin{itemize}
\item Darkode:
Cybercriminal wares, including exploit kits, spam services,
ransomware programs, and stealthy botnets.
\item Hack Forums:
A mixture of cyber-security and computer gaming blackhat and non-cybercrime products.
\item Blackhat: 
Blackhat Search Engine Optimization techniques.
\item Nulled: 
Data stealing tools and services.
\end{itemize}

Table~\ref{table:datasets} gives some statistics of these forums. These are the same
forums used to study product activity in \newcite{PortnoffEtAl2017}.
We collected all available posts and annotated a subset of them.
In total, we annotated 130,336 tokens; accounting for multiple annotators, our annotators considered 478,176 tokens in the process of labeling the data.

Figure~\ref{fig:example} shows two examples of posts from Darkode. In
addition to aspects of the annotation, which we describe below, we
see that the text exhibits common features of web text: abbreviations,
ungrammaticality, spelling errors, and visual formatting, particularly in
thread titles. Also, note how some words that are not products here might
be in other contexts (e.g.,~\emph{Exploits}).

\subsection{Annotation Process}

We developed our annotation guidelines through six preliminary rounds of annotation, covering 560 posts.
Each round was followed by discussion and resolution of every post with disagreements.
We benefited from members of our team who brought extensive domain expertise to the task.
As well as refining the annotation guidelines, the development process trained annotators who were not security experts.  
The data annotated during this process is not included in Table~\ref{table:datasets}.

Once we had defined the annotation standard, we annotated datasets from Darkode, Hack Forums, Blackhat, and Nulled as described in Table~\ref{table:datasets}.\footnote{The table does not include additional posts that were labeled by all annotators in order to check agreement.}
Three people annotated every post in the Darkode training, Hack Forums training, Blackhat test, and Nulled test sets; these annotations were then merged into a final annotation by majority vote.
The development and test sets for Darkode and Hack Forums were annotated by additional team members (five for Darkode, one for Hack Forums), and then every disagreement was discussed and resolved to produce a final annotation.
The authors, who are researchers in either NLP or computer security, did all of the annotation.

We preprocessed the data using the tokenizer and sentence-splitter from the Stanford CoreNLP toolkit \cite{ManningEtAl2014}.
Note that many sentences in the data are already delimited by line breaks, making the sentence-splitting
task much easier.
We performed annotation on the tokenized data so that annotations would be consistent with
surrounding punctuation and hyphenated words.

Our full annotation guide is available with our data release.\footnote{\texttt{https://evidencebasedsecurity.org/\\forums/annotation-guide.pdf}} Our
basic annotation principle is to annotate tokens when they are either the
product that will be delivered or are an integral part of the method leading
to the delivery of that product. Figure~\ref{fig:example} shows examples of
this for a deliverable product (\emph{bot}) as well as a service (\emph{cleaning}).
Both a product and service may be annotated in a single example: for a post asking
to \emph{hack an account}, \emph{hack} is the method and the deliverable
is the \emph{account}, so both are annotated.
In general, methods expressed as verbs may be annotated in addition to
nominal references.

When the product is a multiword expression (e.g.,~\emph{Backconnect bot}), it is almost exclusively a noun phrase, in which case we annotate the head word of the noun phrase (\emph{bot}).
Annotating single tokens instead of spans meant that we avoided having to agree on an exact parse of each post, since even the boundaries of base noun phrases can be quite difficult to agree on in ungrammatical text.

If multiple different products are being bought or sold, we annotate them all.
We do not annotate:
\begin{itemize}
    \item Features of products
    \item Generic product references, e.g.,~\emph{this}, \emph{them}
    \item Product mentions inside ``vouches'' (reviews from other users)
    \item Product mentions outside of the first and last 10 lines of each
      post\footnote{In preliminary annotation we found that content in the middle of the post
      typically described features or gave instructions without explicitly mentioning
      the product.  Most posts are unaffected by this rule: 96\% of Darkode,
      77\% of Hack Forums, 84\% of Blackhat, and 93\% of Nulled posts are less than 20 lines. However, the cutoff still substantially
      reduced annotator effort on the tail of very long posts.}
\end{itemize}

Table~\ref{table:datasets} shows inter-annotator agreement according to our
annotation scheme.  We use the Fleiss' Kappa measurement \cite{Kappa},
treating our task as a token-level annotation where every token is annotated
as either a product or not.  We chose this measure as we are interested
in agreement between more than two annotators (ruling out Cohen's kappa),
have a binary assignment (ruling out correlation coefficients) and have
datasets large enough that the biases Krippendorff's Alpha addresses are
not a concern.  The values indicate reasonable agreement.

\subsection{Discussion}

Because we annotate entities in a context-sensitive way (i.e.,~only
annotating those in product context), our task resembles a post-level
information extraction task. The product information in a post can be thought of
as a list-valued slot to be filled in the style of TAC
KBP \cite{Surdeanu2013,SurdeanuJi2014}, with the token-level annotations
constituting provenance information. However, we chose to anchor the task
fully at the token level to simplify the annotation task:
at the post level, we would have to decide whether two distinct product
mentions were actually distinct products or not, which requires heavier
domain knowledge. Our approach also resembles the fully token-level
annotations of entity and event information in the ACE dataset \cite{ACE2005}.

\section{Evaluation Metrics}
\label{sec:evaluation}

In light of the various views on this task and its different requirements for
different potential applications, we describe and motivate a few distinct
evaluation metrics below. The choice of metric will impact system design,
as we discuss in the following sections.

\paragraph{Token-level accuracy}
We can follow the approach used in token-level tasks like NER and
compute precision, recall, and F$_1$ over the set of tokens labeled as
products. This most closely mimics our annotation process.

\paragraph{Type-level product extraction (per post)}
For many applications, the primary goal of the extraction task is more in
line with KBP-style slot filling, where we care about the set of products
extracted from a particular post. Without a domain-specific lexicon
containing full synsets of products (e.g., something that
could recognize that \emph{hack} and \emph{access} are synonymous),
it is difficult to evaluate this in a fully satisfying way.
However, we approximate this evaluation by comparing the set of product \emph{types}\footnote{Two product tokens are considered the same type if after lowercasing and stemming they have a sufficiently small edit distance: 0 if the tokens are length 4 or less, 1 if the lengths are between 5 and 7, and 2 for lengths of 8 or more} in a post with the set of product types predicted by the system. Again, we consider precision, recall,
and F$_1$ over these two sets. This metric favors systems that consistently make correct
post-level predictions even if they do not retrieve every token-level occurrence of the product.

\paragraph{Post-level accuracy}
Most posts contain only one product, but our type-level extraction will naturally be a conservative estimate of
performance simply because there may seem to be multiple ``products'' that
are actually just different ways of referring to one core product.
Roughly 60\% of posts in the two forums contain multiple annotated tokens
that are distinct beyond stemming and lowercasing.  However, we analyzed
100 of these multiple product posts across Darkode and Hack Forums, and
found that only 6 of them were actually selling multiple products, indicating
that posts selling multiple types of products are actually quite rare
(roughly 3\% of cases overall).
In the rest of the cases, the variations
were due to slightly different ways of describing the same product.

In light of this, we also might consider asking the system to extract
\emph{some} product reference from the post, rather than all of them.
Specifically, we compute accuracy on a post-level by checking whether the first product type
extracted by the system is contained in the annotated
set of product types.\footnote{For this metric we exclude posts containing no products. These are usually posts that have had their content deleted or are about forum administration.} Because most posts feature one product, this metric
is sufficient to evaluate whether we understood what the core product of the post was.

\subsection{Phrase-level Evaluation} \label{sec:NP-eval}

Another axis of variation in metrics comes from whether we consider
token-level or phrase-level outputs.  As noted in the previous section,
we did not annotate noun phrases, but we may actually be interested in identifying them.
In Figure~\ref{fig:example}, for example, extracting 
\emph{Backconnect bot} is more useful than extracting \emph{bot} in isolation,
since \emph{bot} is a less specific characterization of the product.

We can convert our token-level annotations to phrase-level annotations
by projecting our annotations to the noun phrase level based on the output
of an automatic parser. We used the parser of
\newcite{ChenManning2014} to parse all sentences of each post. For each
annotated token that was given a nominal tag (N*), we projected that token
to the largest NP containing it of length less than or equal to 7; most
product NPs are shorter than this, and when the parser predicts a longer
NP, our analysis found that
it typically reflects a mistake. In Figure~\ref{fig:example}, the entire noun
phrase \emph{Backconnect bot} would be labeled as a product. For
products realized as verbs (e.g., \emph{hack}), we leave the annotation as
the single token. \\

Throughout the rest of this work, we will evaluate sometimes at the
token-level and sometimes at the NP-level\footnote{Where NP-level means
``noun phrases and verbs'' as described in Section~\ref{sec:NP-eval}.} (including for the product type
evaluation and post-level accuracy); we will specify which evaluation is
used where.

\section{Models}
\label{sec:models}

We consider several baselines for product extraction,
two supervised learning-based methods (here), and semi-supervised methods (Section~\ref{sec:domain-adapt}).

\paragraph{Baselines} One approach takes the most \textbf{frequent} noun
or verb in a post and classifies all occurrences of that word type as products.
A more sophisticated lexical baseline is based on a product \textbf{dictionary} extracted
from our training data: we tag the most frequent noun or verb in a post
that also appears in this dictionary. This method
fails primarily in that it prefers to extract common words like \emph{account}
and \emph{website} even when they do not occur as products.
The most relevant off-the-shelf system is an \textbf{NER} tagging model; we retrain the Stanford NER system on our data \cite{stanford-ner}.
Finally, we can
tag the \textbf{first} noun phrase of the post as a product, which will often capture
the product if it is mentioned in the title of the post.\footnote{Since this baseline
fundamentally relies on noun phrases, we only evaluate it in the noun phrase setting.}

We also include human performance results.
We averaged the results for annotators compared with the consensus annotations.
For the phrase level evaluation, we apply the projection method described in Section~\ref{sec:NP-eval}.

\paragraph{Binary classifier/CRF} One learning-based approach to this task is to
employ a binary SVM classifier for each token in isolation. We also experimented
with a token-level CRF with a binary tagset, and found identical performance, so we describe
the binary classifier version.\footnote{We further experimented with a bidirectional LSTM tagger and found similar performance as well.} Our features look at both the token under
consideration as well as neighboring tokens, as described in the next paragraph.
A vector of ``base features'' is extracted for each of these target
tokens: these include 1) sentence position in
the document and word position in the current sentence as bucketed indices; 2) word identity
(for common words), POS tag, and dependency relation to parent for each word
in a window of size 3 surrounding the current word; 3) character 3-grams
of the current word. The same base feature set is used for every token.

Our token-classifying SVM extracts base features on the token under
consideration as well as its syntactic parent. Before inclusion in the final classifier, these features are
conjoined with an indicator of their source (i.e., the current token or the parent token).
Our NP-classifying SVM extracts base features on first, last, head, and syntactic parent
tokens of the noun phrase, again with each feature conjoined with its token source.

We weight false positives and false negatives differently to adjust the precision/recall
curve (tuned on development data for each forum), and we also empirically
found better performance by upweighting the contribution to the objective
of singleton products (product types that occur only once in the training
set).

\paragraph{Post-level classifier}
As discussed in Section~\ref{sec:evaluation},
one metric we are interested in is whether we can find \emph{any} occurrence of
a product in a post. This task is easier than the general tagging problem: if we can effectively identify the
product in, e.g., the title of a post, then we do not need to identify
additional references to that product in the body of the post. Therefore,
we also consider a post-level model, which directly tries to select one token (or
NP) out of a post as the most likely product. Structuring the prediction
problem in this way naturally lets the model be more conservative in its
extractions, since highly ambiguous product mentions can be
ignored if a clear product mention is present. Put another way, it supplies
a useful form of prior knowledge, namely that each post has exactly one product in almost all cases.

Our post-level system is formulated as an instance of a latent SVM \cite{YuJoachims2009}.
The output space is the set of all tokens (or noun phrases, in the NP case) in the post.
The latent variable is the choice of token/NP to select, since there may be
multiple correct choices of product tokens. The features used on each token/NP are the same
as in the token classifier.

We trained all of the learned models by subgradient descent on the primal form
of the objective \cite{RatliffEtAl2007,KummerfeldEtAl2015}. We use AdaGrad
\cite{DuchiEtAl2011} to speed convergence in the presence of a large weight
vector with heterogeneous feature types. All product extractors in this
section are trained for 5 iterations with $\ell_1$-regularization tuned
on the development set.

\subsection{Basic Results}
\begin{table}[t]
\small
\begin{center}
\renewcommand{\tabcolsep}{1.6mm}
\begin{tabular}{rccc|ccc|c} \hline
\multicolumn{8}{c}{Token Prediction} \\
 & \multicolumn{3}{c}{Tokens} & \multicolumn{3}{c}{Products} & Posts \\
 & P & R & F$_1$ & P & R & F$_1$ & Acc. \\  \hline
Freq   & 41.9 & 42.5 & 42.2 & 48.4 & 33.5 & 39.6 & 45.3 \\
Dict   & 57.9 & 51.1 & 54.3 & 65.6 & 44.0 & 52.7 & 60.8 \\ 
NER    & 59.7 & 62.2 & 60.9 & 60.8 & 62.6 & 61.7 & 72.2 \\ \hline
Binary & 62.4 & 76.0 & \textbf{68.5} & 58.1 & 77.6 & 66.4 & 75.2 \\
Post   & 82.4 & 36.1 & 50.3 & 83.5 & 56.6 & 67.5 & \textbf{82.4} \\ \hline\hline
Human$^*$& 86.9 & 80.4 & 83.5 & 87.7 & 77.6 & 82.2 & 89.2 \\ \hline
\multicolumn{8}{c}{NP Prediction} \\
 & \multicolumn{3}{c}{NPs} & \multicolumn{3}{c}{Products} & Posts \\
 & P & R & F$_1$ & P & R & F$_1$ & Acc. \\ \hline
Freq   & 61.8 & 28.9 & 39.4 & 61.8 & 50.0 & 55.2 & 61.8 \\
Dict   & 57.9 & 61.8 & 59.8 & 71.8 & 57.5 & 63.8 & 68.0 \\
First  & 73.1 & 34.2 & 46.7 & 73.1 & 59.1 & 65.4 & 73.1 \\
NER & 63.6 & 63.3 & 63.4 & 69.7 & 70.3 & 70.0 & 76.3 \\ \hline
Binary & 67.0 & 74.8 & \textbf{70.7} & 65.5 & 82.5 & 73.0 & 82.4 \\
Post   & 87.6 & 41.0 & 55.9 & 87.6 & 70.8 & \textbf{78.3} & \textbf{87.6} \\ \hline\hline
Human$^*$ & 87.6 & 83.2 & 85.3 & 91.6 & 84.9 & 88.1 & 93.0 \\ \hline
\end{tabular}
\end{center}
\caption{\label{table:baseline} Development set results on Darkode. Bolded
	F$_1$ values represent
	statistically-significant improvements over all other system values in
	the column with $p < 0.05$ according to a bootstrap resampling
	test. Our post-level system outperforms our binary classifier
	at whole-post accuracy and on type-level product extraction, even though
	it is less good on the token-level metric. All systems consistently identify product NPs
better than they identify product tokens. However, there is a substantial gap between our systems and human performance.}
\end{table}

Table~\ref{table:baseline} shows development set results on Darkode for
each of the four systems for each metric described in
Section~\ref{sec:evaluation}. Our learning-based systems substantially
outperform the baselines on the metrics they are optimized for. The
post-level system underperforms the binary classifier on the token
evaluation, but is superior at not only post-level accuracy but also
product type F$_1$. This lends
credence to our hypothesis that picking one product suffices to
characterize a large fraction of posts. Comparing the automatic systems with human annotator
performance we see a substantial gap.
Note that our best annotator's token F$_1$ was 89.8, and NP post accuracy was 100\%;
a careful, well-trained annotator can achieve
very high performance, indicating a high skyline.

The noun phrase metric appears to be generally
more forgiving,
since token distinctions within noun phrases are erased.
The post-level NP system achieves an F-score of 78 on product type identification,
and post-level accuracy is around 88\%. While there is room
for improvement, this system is accurate enough to enable
analysis of Darkode with automatic annotation.

Throughout the rest of this work, we focus on NP-level evaluation and post-level
NP accuracy.

\section{Domain Adaptation} \label{sec:domain-adapt}

Table~\ref{table:baseline} only showed results for training and evaluating within
the same forum (Darkode). However, we wish to apply
our system to extract product occurrences from a wide variety of forums,
so we are interested in how well the system will generalize to a new forum.
Tables~\ref{table:domaintok} and~\ref{table:domainpost} show full
results of several systems in within-forum and cross-forum evaluation
settings. Performance is severely degraded in the cross-forum setting
compared to the within-forum setting, e.g.,~on NP-level F$_1$, a Hack Forums-trained model is 14.6
F$_1$ worse at the Darkode task than a Darkode-trained model (61.2 vs.~75.8). Differences in how the systems adapt
between different forums will be explored more thoroughly in
Section~\ref{sec:analysis}.

In the next few sections, we explore several
possible methods for improving results in the cross-forum settings and
attempting to build a more domain-general system. These techniques generally reflect
two possible hypotheses about the source of the cross-domain challenges:

\paragraph{Hypothesis 1:} Product inventories are the primary difference across
domains; context-based features will transfer, but the main challenge is not
being able to recognize unknown products.

\paragraph{Hypothesis 2:} Product inventories \textbf{and} stylistic conventions both differ across
domains; we need to capture both to adapt models successfully.

\begin{table*}[t]
\begin{center}
\small
  \setlength{\tabcolsep}{5.5pt}
\begin{tabular}{l|rrr|rrr|rrr|rrr|r} \hline
  \multirow{2}{*}{\diagbox{System\phantom{......}}{Eval data}} & \multicolumn{3}{c}{Darkode} & \multicolumn{3}{|c}{Hack Forums} & \multicolumn{3}{|c}{Blackhat} & \multicolumn{3}{|c|}{Nulled} & Avg \\
  & P & R & F$_1$ & P & R & F$_1$ & P & R & F$_1$ & P & R & F$_1$ & F$_1$  \\ \hline
 \multicolumn{13}{c}{Trained on Darkode} \\ \hline
Dict                    & 55.9 & 54.2 & \phantom{$\dagger$}55.0 & 42.1 & 39.8 & 40.9 & 37.1 & 36.6 & 36.8 & 52.6 & 43.2 & 47.4 & 45.0 \\ \hline
Binary                  & 73.3 & 78.6 & \phantom{$\dagger$}75.8          & 51.1 & 50.2 & 50.6          & 55.2 & 58.3 & 56.7 & 55.2 & 64.0 & 59.3 & 60.6 \\
 Binary + Brown Clusters & 75.5 & 76.4 & \phantom{$\dagger$}\textbf{76.0} & 55.9 & 48.1 & 51.7          & 59.7 & 57.1 & \textbf{58.4} & 60.0 & 61.1 & \textbf{60.5} & 61.7 \\
Binary + Gazetteers     & 73.1 & 75.6 & \phantom{$\dagger$}74.3          & 52.6 & 51.1 & \textbf{51.8} & $-$\phantom{-} & $-$\phantom{-} & $-$\phantom{-} & $-$\phantom{-} & $-$\phantom{-} & $-$\phantom{-} & $-$\phantom{-} \\ \hline
 \multicolumn{13}{c}{Trained on Hack Forums} \\ \hline
Dict                    & 57.3 & 44.8 & \phantom{$\dagger$}50.3 & 50.0 & 52.7 & 51.3 & 45.0 & 44.7 & 44.8 & 51.1 & 43.6 & 47.1 & 48.4 \\ \hline
Binary                  & 67.0 & 56.4 & \phantom{$\dagger$}61.2 & 58.0 & 64.2 & 61.0          & 62.4 & 60.8 & \textbf{61.6} & 71.0 & 68.9 & 69.9 & 63.4 \\
Binary + Brown Clusters & 67.2 & 52.5 & \phantom{$\dagger$}58.9 & 59.3 & 64.7 & \textbf{61.9} & 61.9 & 59.6 & 60.7 & 73.1 & 67.4 & \textbf{70.2} & 62.9 \\
Binary + Gazetteers      & 67.8 & 64.1 & $\dagger$\textbf{65.9}  & 59.9 & 61.3 & 60.6 & $-$\phantom{-} & $-$\phantom{-} & $-$\phantom{-} & $-$\phantom{-} & $-$\phantom{-} & $-$\phantom{-} & $-$\phantom{-} \\ \hline
\end{tabular}
\vspace{-0.1in}
\end{center}
\caption{\label{table:domaintok} Test set results at the NP level
	in within-forum and cross-forum settings for a variety of different
	systems.  Using either Brown clusters or gazetteers gives mixed
	results on cross-forum performance: only one of the improvements ($\dagger$) is
	statistically significant with $p < 0.05$ according to a bootstrap
	resampling test. Gazetteers are unavailable for Blackhat and Nulled since we have no training data for those forums.}
\vspace{-0.1in}
\end{table*}

\subsection{Brown Clusters}

To test Hypothesis 1, we investigate whether additional lexical information
helps identify product-like words in new domains.
A classic semi-supervised technique for exploiting unlabeled target data
is to fire features over word clusters
or word vectors \cite{TurianEtAl2010}. These features should generalize
well across domains that the clusters are formed on: if product
nouns occur in similar contexts across domains and therefore wind up in the same
cluster, then a model trained on domain-limited data should be able
to learn that that cluster identity is indicative of products.

We form Brown clusters on our unlabeled data from both Darkode and Hack Forums
(see Table~\ref{table:datasets} for sizes).  We use \newcite{Liang2005}'s 
implementation to learn 50 clusters.\footnote{This value was chosen based on dev set experiments.} Upon inspection, these clusters do indeed capture some of the
semantics relevant to the problem: for example, the cluster 110 has
as its most frequent members \emph{service}, \emph{account}, \emph{price},
\emph{time}, \emph{crypter}, and \emph{server}, many of which are
product-associated nouns. We incorporate these as features into our
model by characterizing each token with prefixes of the Brown cluster ID;
we used prefixes of length 2, 4, and 6.

Tables~\ref{table:domaintok} and~\ref{table:domainpost} show the
results of incorporating Brown cluster features into our trained models.
These features do not lead to statistically-significant gains in either
NP-level F$_1$ or post-level accuracy, despite small improvements in some cases.
This indicates that Brown clusters might be a useful feature sometimes, but do not solve the domain adaptation
problem in this context.\footnote{We could also use vector representations of words here, but in initial experiments, these did not outperform Brown clusters. That is consistent with the results of \newcite{TurianEtAl2010} who showed
similar performance between Brown clusters and word vectors for chunking and NER.}

\subsection{Type-level Annotation}
\label{sec:type-level}

Another approach following Hypothesis 1 is to use small amounts of supervised data,
One cheap approach
for annotating data in a new domain is to exploit type-level annotation
\cite{GarretteBaldridge2013,GarretteEtAl2013}. Our token-level annotation
standard is relatively complex to learn, but a researcher could
quite easily provide a few exemplar products for a new forum based on just
a few minutes of reading posts and analyzing the forum.

\begin{table}[t]
\begin{center}
\small
  \setlength{\tabcolsep}{5pt}
\begin{tabular}{rcccc} \hline
  & Darkode & Hack Forums & Blackhat & Nulled \\ \hline
 \multicolumn{5}{c}{Trained on Darkode} \\ \hline
Dict   & \phantom{$\dagger$}59.3 & \phantom{$\dagger$}39.7 & 43.5 & 54.6 \\ \hline
Post   & \phantom{$\dagger$}\textbf{89.5} & \phantom{$\dagger$}66.9 & \textbf{75.8} & 79.0\\
+Brown & \phantom{$\dagger$}\textbf{89.5} & \phantom{$\dagger$}66.9 & 69.3 & \textbf{84.8}\\
+Gaz   & \phantom{$\dagger$}87.5 & \phantom{$\dagger$}\textbf{72.1} & $-$ & $-$ \\ \hline
 \multicolumn{5}{c}{Trained on Hack Forums} \\ \hline
Dict   & \phantom{$\dagger$}48.9 & \phantom{$\dagger$}53.6 & 50.0 & 53.4 \\ \hline
Post   & \phantom{$\dagger$}78.1 & \phantom{$\dagger$}78.6 & 74.1 & 81.3 \\ 
+Brown & \phantom{$\dagger$}\textbf{82.2} & \phantom{$\dagger$}81.6 & \textbf{77.4} & \textbf{82.5} \\
+Gaz   & \phantom{$\dagger$}79.1 & $\dagger$\textbf{83.8} & $-$ & $-$ \\ \hline
\end{tabular}
\end{center}
\caption{\label{table:domainpost} Test set results at the whole-post level
	in within-forum and cross-forum settings for a variety of different
	systems.  Brown clusters and gazetteers give similarly mixed results as in the
	token-level evaluation; $\dagger$ indicates statistically significant gains over
	the post-level system with $p < 0.05$ according to a bootstrap
	resampling test.}
\end{table}

Given the data that we've already annotated, we can simulate this
process by iterating through our labeled data and collecting annotated
product names that are sufficiently common. Specifically, we take all
(lowercased, stemmed) product tokens and keep those occurring at least 4
times in the training dataset (recall that these datasets are $\approx$~700
posts). This gives us a list of 121 products in Darkode and 105 products
in Hack Forums.

\begin{table*}[t]
\small
  \begin{center}
    \setlength{\tabcolsep}{4pt}
  \begin{tabular}{l|ccc|ccc|ccc|ccc}
    \hline
    \multirow{2}{*}{\diagbox{System}{Test}} & \multicolumn{3}{c|}{Darkode} & \multicolumn{3}{c|}{Hack Forums} & \multicolumn{3}{c|}{Blackhat} & \multicolumn{3}{c}{Nulled} \\ 
            & \% OOV & R$_\textrm{seen}$ & R$_\textrm{oov}$ & \% OOV & R$_\textrm{seen}$ & R$_\textrm{oov}$ & \% OOV & R$_\textrm{seen}$ & R$_\textrm{oov}$ & \% OOV & R$_\textrm{seen}$ & R$_\textrm{oov}$ \\ \hline
Binary (Darkode)     & 20 & 78 & 62 & 41 & 64 & 47 & 42 & 69 & 46 & 30 & 72 & 45 \\
Binary (HF)          & 50 & 76 & 40 & 35 & 75 & 42 & 51 & 70 & 38 & 33 & 83 & 32 \\ 
    \hline
\end{tabular}
\end{center}
\caption{\label{table:oov} Product token out-of-vocabulary rates on development sets
	(test set for Blackhat and Nulled) of various forums with respect to training
	on Darkode and Hack Forums. We also show the recall of an NP-level
	system on seen (R$_\textrm{seen}$) and OOV (R$_\textrm{oov}$) tokens. Darkode seems to be more ``general''
	than Hack Forums: the Darkode system generally has lower OOV rates
	and provides more consistent performance
	on OOV tokens than the Hack Forums system.
}
\end{table*}

To incorporate this information into our system, we add a new feature on
each token indicating whether or not it occurs in the gazetteer. At training
time, we use the gazetteer scraped from the training set. At test time,
we use the gazetteer from the target domain as a form of partial type-level supervision.
Tables~\ref{table:domaintok} and \ref{table:domainpost} shows the results
of incorporating the gazetteer into the system. Gazetteers seem to
provide somewhat consistent gains in cross-domain settings, though many of these
individual improvements are not statistically significant, and the gazetteers
can sometimes hurt performance when testing on the same domain the system was trained on.

\subsection{Token-level Annotation}
\label{sec:token-level}

We now turn our attention to methods that might address Hypothesis 2.
If we assume the domain transfer problem is more complex, we really want
to leverage labeled data in the target domain rather than attempting to transfer features based only on
type-level information. Specifically, we are interested in cases where a relatively
small number of labeled posts (less than 100) might provide substantial benefit
to the adaptation; a researcher could plausibly do this annotation
in a few hours.

\begin{figure}[t!]
\begin{centering}
\includegraphics[trim=0mm 108mm 00mm 0mm,scale=0.52]{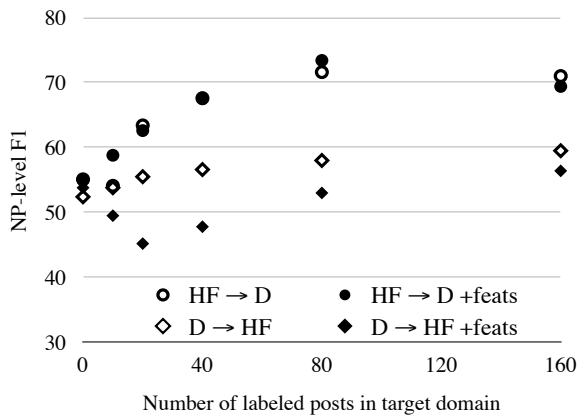}
\caption{\label{fig:plot-adapt} Token-supervised domain adaptation results
	for two settings. As our system is trained on an increasing amount
	of target-domain data (x-axis), its performance generally improves.
	However, adaptation from Hack Forums to Darkode is much more effective
	than the other way around, and using domain features
	as in \protect\newcite{Daume2007} gives little benefit over na\"{i}ve
	use of the new data.
}
\end{centering}
\end{figure}

We consider two ways of exploiting labeled target-domain
data. The first is to simply take these posts as additional training data.
The second is to also employ the ``frustratingly easy'' domain
adaptation method of \newcite{Daume2007}. In this framework, each feature fired
in our model is actually fired twice: one copy is domain-general and one is conjoined
with the domain label (here, the name of the forum).\footnote{If we are training on data
from $k$ domains, this gives rise to up to $k+1$ total versions of each feature.}
In doing so, the model should gain
some ability to separate domain-general from domain-specific feature values, with regularization
encouraging the domain-general feature to explain as much of the phenomenon as possible.
For both training methods, we upweight the contribution of the target-domain posts in
the objective by a factor of 5.

Figure~\ref{fig:plot-adapt} shows learning curves for both of these methods
in two adaptation settings as we vary the amount of labeled target-domain
data. The system trained on Hack Forums is able to make good use of labeled
data from Darkode: having access to 20 labeled posts leads to gains of
roughly 7 F$_1$. Interestingly, the system trained on Darkode is not able
to make good use of labeled data from Hack Forums, and the domain-specific
features actually cause a drop in performance until we
include a substantial amount of data from Hack Forums (at least 80 posts). We are likely overfitting the small Hack Forums training set with the domain-specific features.

\subsection{Analysis}
\label{sec:analysis}

In order to understand the variable performance and shortcomings of the
domain adaptation approaches we explored, it is useful to examine our
two initial hypotheses and characterize the datasets a bit further. To do so, we
break down system performance on products seen in the training set versus novel
products. Because our systems depend on lexical and character $n$-gram features,
we expect that they will do better at predicting products we have seen before.

Table~\ref{table:oov} confirms this intuition: it shows product out-of-vocabulary
rates in each of the four forums relative to training on both Darkode and
Hack Forums, along with recall of an NP-level system on both previously
seen and OOV products. As expected, performance is substantially higher
on in-vocabulary products. OOV rates of a Darkode-trained system are generally lower on
new forums, indicating that that forum has better all-around product coverage.
A system trained on Darkode is therefore in some sense
more domain-general than one trained on Hack Forums. 

This would seem to support Hypothesis 1. Moreover, Table~\ref{table:domaintok} shows that
the Hack Forums-trained system achieves a 21\% error reduction on Hack Forums compared
to a Darkode-trained system, while a Darkode-trained system
obtains a 38\% error reduction on Darkode relative to a Hack Forums-trained system; this greater
error reduction means that Darkode has better coverage of Hack Forums than vice versa.
Darkode's better product coverage
also helps explain why Section~\ref{sec:token-level}
showed better performance of adapting Hack Forums to Darkode than the other way around:
augmenting Hack Forums data with a few posts from Darkode can give critical knowledge about
new products, but this is less true if the forums are reversed. Duplicating features and adding parameters to the
learner also has less of a clear benefit when adapting from Darkode, when the
types of knowledge that need to be added are less concrete.

Note, however, that these results do not tell the full story. Table~\ref{table:oov} reports recall values,
but not all systems have the same precision/recall tradeoff: although they were tuned
to balance precision and recall on their respective development sets, the Hack Forums-trained
system is slightly more precision-oriented on Nulled than the Darkode-trained system.\footnote{While a hyperparameter controlling the precision/recall tradeoff could theoretically be tuned on the target domain, it is hard to do this in a robust, principled way without having access to a sizable annotated dataset from that domain. This limitation further complicates the evaluation and makes it difficult to set up apples-to-apples comparisons across domains.} In fact, Table~\ref{table:domaintok}
shows that the Hack Forums-trained system actually performs better on Nulled,
largely due to better performance on previously-seen products.
This indicates that there is some truth to Hypothesis 2: product coverage is not
the only important factor determining performance.

\section{Conclusion}

We present a new dataset of posts from cybercrime marketplaces annotated with
product references, a task which blends IE and NER.
Learning-based methods degrade in performance when applied to new forums, and while we explore methods for
fine-grained domain adaption in this data, effective methods for this task are
still an open question.

Our datasets used in this work are available at
{\fontsize{8.9}{10.9} \url{https://evidencebasedsecurity.org/forums/} }
Code for the product extractor can be found at
{\fontsize{8.9}{10.9} \url{https://github.com/ccied/ugforum-analysis/tree/master/extract-product}}

\section*{Acknowledgments}

This work was supported in part by the National Science Foundation under grants CNS-1237265 and CNS-1619620, by the Office of Naval Research under MURI grant N000140911081, by the Center for Long-Term Cybersecurity and by gifts from Google. We thank all the people that provided us with forum data for our analysis; in particular Scraping Hub and SRI for their assistance in collecting data for this study. Any opinions, findings, and conclusions expressed in this material are those of the authors and do not necessarily reflect the views of the sponsors.

\bibliography{forum-emnlp2017}
\bibliographystyle{emnlp_natbib}

\end{document}


\maketitle

\section{General Guidelines}
Annotate every point in the post when the product is referred to explicitly.
In each case, label the single word that best captures its meaning.

\begin{itemize}
  \item To indicate the word, enclose it in:\footnote{For this work we do not use the buy/sell distinction that is included in the annotations.}
    \begin{itemize}
      \item If it is being sold, use '\{' and '\}'
      \item If it is being bought, use '[' and ']'
    \end{itemize}
    or
    \begin{itemize}
      \item If it is being sold, use '\{S' and '\}'
      \item If it is being bought, use '\{B' and '\}'
    \end{itemize}
  \item Braces should enclose one whitespace delimited token.
  \item Only annotate cases in the first 10 lines and last 10 lines of a file (not counting blank lines). The numbers at the start of lines provide the relevant count.
  \item Do not annotate text inside vouches, marked by lines saying <blockquote> (but vouches do count towards the 10 line rule).
  \item If there are multiple products, annotate all of them. However, do not annotate features of a single product. \\
     e.g. do not label \textit{Logo} in \textit{Logo for the [app] 5 \$}
  \item If a product is referred to with a generic word then do not annotate it. \\
     e.g. \textit{somebody}, \textit{it}, \textit{get} \\
   The exception to this rule is if the product is never referred to more specifically, then do annotate. \\
     e.g. \textit{a [method] for making money}
  \item If the product is a service that produces an item, annotate both. \\
     e.g. \textit{[hack] a big [website]}, and \textit{I want a coder to [write] an [exploit]}
  \item Do annotate specific product names if that is the way they are mentioned, \\
     e.g. \textit{[FaceSpread] / / Custom Panel / / HTTP crypter and more !}
  \item Do not annotate unrelated or rival products that are mentioned \\
     e.g. \textit{similar to Element Scanner or other scanning websites}
\end{itemize}

\paragraph{Multi-word expressions}
Here we are referring to cases like \textit{Remote Administration Tool}, where there are a series of consecutive words forming a single mention of the product.
Select only one word, trying to choose the one that best captures the type of product.
Some examples of common cases (with recommended annotations) are:

\begin{itemize}
  \item \textit{Facebook [accounts]} \\
    \textit{Facebook} would be less specific, as it could occur with accounts, likes,
    hacking of the company itself, etc.
  \item \textit{Phone [Verification] Service} \\
    Here \textit{Service} would be the next best choice, but \textit{verification} is a
    better characterization.
  \item \textit{YouTube Multi Account [Subscriber] Bot} \\
    Similar idea to the previous case.
\end{itemize}

One heuristic to try is thinking about what words can and cannot be dropped and
still have the product be the same. Those that cannot be dropped are better
options for annotation.

\subsection{Flags}

If a post falls into one of the categories below, we add an extra line at the end of the file and write the letter for that case:\footnote{These flags were primarily included for exploratory analysis and characterization. They are not used in any way in the work described in the main paper.}

\begin{description}
  \item[A] ADMIN posts, about running the site, deleting a thread, etc
  \item[D] DIFFICULT cases
  \item[W] WEIRD cases
  \item[G] GAMING related posts
  \item[L] Posts that are not interesting to us (e.g. selling signature space, or shoes)
\end{description}

\section{Tricky example cases for reference}

Complex products:
\begin{itemize}
  \item \textit{i am looking to buy USA \{Dob\} + \{ssn\} + \{fname\} + \{lname\} + \{address\}}
  \item \textit{I am buying as many \{email\} : \{passes\} as possible.}
  \item \textit{\{Adbot\}-\{clickbot\} \{coder\} needed.}
  \item \textit{\{WMZ\} - [LR] \\
  1:1 \\
  Around 200\$}
\end{itemize}

Combined items and services:
\begin{itemize}
  \item \textit{Steal certain @yahoo \{email\} or facebook \{account\} \\
    i need a certian @yahoo \{email\} address \{hacked\} or an associated facebook \{account\}, paying good \$\$ if u can do this send me a pm for the info thanx}
  \item \textit{\{hack\} \{website\} \\
    I pay 2000? via WesternUnion \\
    PM or add to MSN if interested.}
\end{itemize}

Different words that refer to the product:
\begin{itemize}
  \item \textit{anyone have AE \{logs\} ? \\
    Pm me , i need some \{links\} will pay 100\$ each}

  \item \textit{MySQLi \{Dumper\} is the best GUI \{tool\} dedicated to SQL injection attacks on MySQL.}

  \item \textit{~250 .de \{rdps\} \\
  \{servers\} and xp, more \{servers\}, some 2008 \{servers\}, some windows 7, good speeds, mostly admin access \\
  mostly have usernames and passwords different than whats on the .ru markets}
\end{itemize}

Multiple products:
\begin{itemize}
  \item \textit{\{SQLi\} and Admin \{Login\} for site with 4.8+ million users}

  \item \textit{need small \{hosting\} space \\
  i need small \{hosting\} space (3-5gb) + one mysql \{base\} to host my php stealer - i wont host exe's or someother shits, only logs. post your price}

  \item \textit{Buy complete c++ sources of svchost \{injection\} and AV \{bypass\} \\
  Hi. \\
  I need you c++ sources - completed, tested and good commented. \\
  Need work and stable AV \{bypass\} code in usermode, for example undetectable svchost process injection.}
\end{itemize}